\documentclass{article}

\PassOptionsToPackage{numbers}{natbib}
\usepackage[preprint]{neurips_2026}

\usepackage[utf8]{inputenc} % allow utf-8 input
\usepackage[T1]{fontenc}    % use 8-bit T1 fonts
\usepackage{hyperref}       % hyperlinks
\usepackage{url}            % simple URL typesetting
\usepackage{booktabs}       % professional-quality tables
\usepackage{amsfonts}       % blackboard math symbols
\usepackage{nicefrac}       % compact symbols for 1/2, etc.
\usepackage{microtype}      % microtypography
\usepackage{xcolor}         % colors
\usepackage{graphicx}
\usepackage{amsmath}
\usepackage{multirow}
\usepackage{makecell}

\title{SpecSem-Net: Integrating Spectral and Semantic Features for Robust AI-generated Video Detection}

\author{%
  Zixi Wei$^1$ \\
  \texttt{zxwei25@stu.pku.edu.cn}
  \And
  Huixuan Zhang$^1$ \\
  \texttt{zhanghuixuan@stu.pku.edu.cn}
  \And
  Xiaojun Wan$^1$\thanks{Corresponding author.} \\
  \texttt{wanxiaojun@pku.edu.cn}
  \AND
  $^1$Wangxuan Institute of Computer Technology, Peking University
}

\begin{document}

\maketitle

\begin{abstract}
The remarkable visual fidelity of recent commercial video generative models, such as Sora and Veo, renders robust AI-generated video detection increasingly essential to prevent synthetic content from being indistinguishable from real videos and exploited for disinformation. However, existing detectors often fail due to an over-reliance on increasingly realistic semantic features, neglecting subtle spectral artifacts. In this paper, we propose \textbf{SpecSem-Net}, the first framework to introduce a semantic-guided spectral denoising mechanism specifically for high-fidelity AI-generated video detection. Specifically, we design a spectral module to extract high-frequency features via Fourier-Transform based filtering. Furthermore, to reduce misjudgments arising from spectral noise, we employ a Gated Merging Mechanism to adaptively fuse semantic context, effectively mitigating spectral noise. Additionally, to evaluate detector performance on the latest top-tier generative models, we construct a comprehensive benchmark comprising 5 SOTA commercial generators. Extensive experiments demonstrate that SpecSem-Net outperforms existing methods, achieving accuracies of \textbf{87.25\%} and \textbf{95.59\%} on our benchmark and public datasets, respectively.
\end{abstract}

% \begin{figure}[t]
%   \centering
%   \includegraphics[width=0.35\textwidth]{figure/radar.pdf} % 建议宽度 0.6-0.7 \textwidth
%   \caption{\textbf{Holistic Performance Overview.} A unified radar chart comparing SpecSem-Net (Ours) against SOTA methods across 12 diverse datasets. The labels are color-coded: \textbf{\textcolor[HTML]{8B0000}{Dark Red}} indicates unseen commercial generators (Wan, Kling, Veo, Sora, Hailuo), while \textbf{\textcolor[HTML]{00008B}{Dark Blue}} indicates open-source benchmarks. Our method (Orange) demonstrates superiority across both domains.}
%   \label{fig:radar_single}
% \end{figure}

\section{Introduction}
The rapid evolution of large-scale video generation models has initiated a new era for content creation. Recent commercial systems, represented by Sora~\cite{openai2025sora2blog}, Kling~\cite{kuaishou2025kling}, and Veo~\cite{deepmind2025veo}, demonstrate superior capabilities of  synthesizing videos characterized by high fidelity and remarkable spatiotemporal consistency. While these technologies significantly empower creative industries, they simultaneously raise the risks of malicious misuse~\cite{liu2025evolvingsinglemodalmultimodalfacial, lee2024tugofwardeepfakegenerationdetection}. If left unchecked, such technologies could be exploited for the mass production of disinformation, severely undermining public trust. Consequently, developing an automated and robust AI-generated video detection mechanism capable of adapting to more and more diverse and powerful generators has become a necessity.

%The increasing visual fidelity of generated videos has rendered explicit artifacts increasingly subtle. Consequently,current detectors' heavy reliance on semantic features overlooks the critical spectral-domain artifacts inherent in the synthesis process, such as\textcolor{red}{add example}. 
Numerous detectors have been developed to address this problem, such as Demamba~\cite{DeMamba} and D3~\cite{zheng2025d3}.
However, these existing detectors tend to rely heavily on artifacts within semantic features of a video, which is actually becoming more and more subtle as video generative models become stronger. Consequently, due to the over-reliance on semantic features, existing detectors suffer from severe performance degradation on videos generated by unseen, state-of-the-art generative models.

While semantic content can be mimicked to deceive the human eyes, spectral-domain artifacts are inherent fingerprints of neural rendering~\cite{frank2020leveraging, durall2020watch}. Further, while spectral artifacts are potential good identifiers of AI-generated videos, pure spectral analysis may struggle to distinguish generative traces from benign high-frequency textures, such as hair or foliage. 

Based on these considerations, we propose \textbf{SpecSem-Net}, a dual-stream framework designed to incorporate spectral features into AI-generated video detection while utilizing semantic features to reduce noise in spectral features. Specifically, we first extract spectral input via Fourier-based filtering and use two separate CLIP-based encoders ~\cite{clip} to achieve semantic features and spectral features correspondingly. At each encoder layer, the current semantic and spectral features are integrated through our \textbf{Gated Merging Mechanism}, which leverages the extracted semantic information to adaptively modulate the spectral features, ensuring a more refined feature. The final extracted spectral feature of each frame are combined and fed into a temporal transformer to achieve a final prediction considering temporal information. 

Furthermore, to provide a more rigorous assessment of how current detectors perform on high-fidelity AI-generated videos, we construct a comprehensive benchmark comprising videos from five cutting-edge commercial video generation models, raising greater challenges to existing detectors. 

In summary, our main contributions are as follows \footnote{Our code and data will be released to the community upon acceptance to facilitate future research.}:

\begin{itemize}
    \item We propose \textbf{SpecSem-Net}, a dual-stream framework that combines spectral and semantic features for AI-generated video detection. 
    
    % \item \textbf{We introduce a targeted temporal sampling strategy that prioritizes the latter segments of a video.} This approach is specifically designed to capture the accumulation of generative traces in later frames, leading to more informative model inputs and improved detection efficiency.
    
    \item We construct a comprehensive benchmark featuring high-fidelity videos from five state-of-the-art commercial generators. This dataset provides a rigorous testing ground for evaluating the generalization capabilities of detectors against the latest generation of synthetic content.
    
    \item Our method achieves superior performance across multiple datasets, outperforming existing detectors by a solid margin. SpecSem-Net reaches an accuracy of \textbf{87.25\%} on our proposed benchmark and \textbf{95.59\%} on the existing GenVideo dataset, demonstrating its effectiveness and strong generalization capability.
\end{itemize}

\begin{figure*}[t]
    \centering
    % 调整宽度：通常设为 \linewidth 或 \textwidth。
    % 如果觉得图太大，可以改为 0.95\linewidth
    \includegraphics[width=1.0\linewidth]{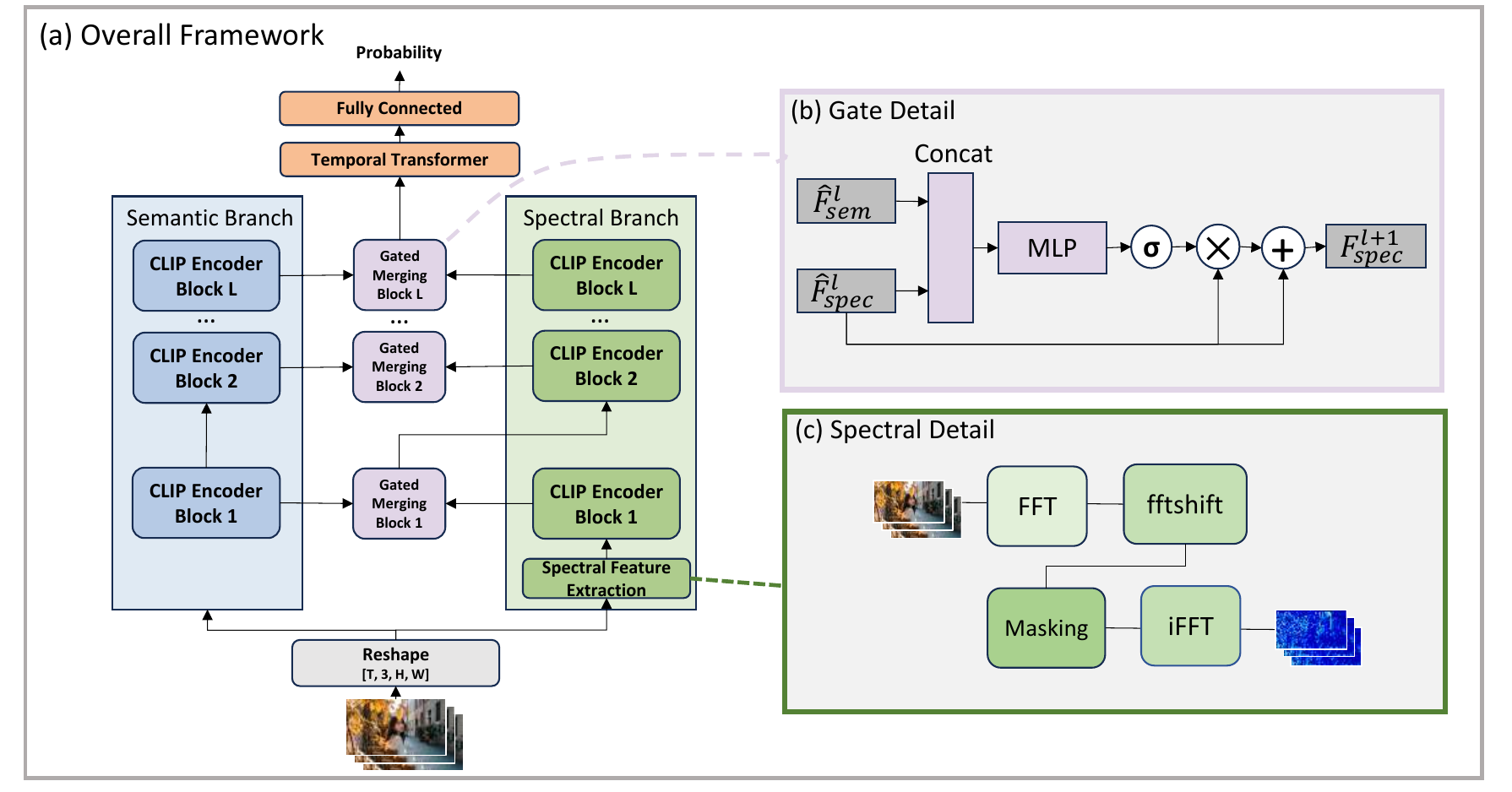} 
    
    \caption{\textbf{Overview of the proposed SpecSem-Net.} 
    (a) The overall dual-stream architecture, comprising a fixed Semantic Branch (Blue) and a trainable Spectral Branch (Green). 
    (b) The \textbf{Gated Merging Mechanism} uses semantic features to dynamically modulate spectral features, filtering out benign environmental noise. 
    (c) The \textbf{Spectral Feature Extraction} module extracts high-frequency residuals via FFT-based high-pass filtering. 
    The dashed lines indicate the detailed composition of the corresponding modules.}
    
    \label{fig:framework}
\end{figure*}

\section{Related Work}
\textbf{AI-Generated Video Detection.}
Deep feature representation constitutes the prevailing paradigm in current research, centering on the utilization of advanced neural architectures to construct discriminative feature spaces for capturing latent visual artifacts. 
Pioneering works primarily targeted semantic inconsistencies, employing pre-trained CNNs or Transformers to extract frame-level representations~\cite{xu2023tall, tiwari2024ai, liu2024turnsimrealrobust, chang2025faraigeneratedvideossimulating}. 
As generative technologies have evolved from early GANs to advanced Diffusion models~\cite{ho2020denoising}, represented by open-source models like Stable Video Diffusion~\cite{blattmann2023stable} and Latte~\cite{ma2024latte}, as well as commercial systems like Gen-3~\cite{runway2024gen3}, Sora~\cite{openai2025sora2blog} and Kling~\cite{kuaishou2025kling}—researchers have transitioned to more potent and diverse backbones~\cite{arnab2021vivit, bertasius2021space, liu2022video} to capture increasingly complex spatiotemporal anomalies.
For instance, recent studies leverage the self-supervised spatiotemporal representation capabilities of VideoMAE~\cite{videomae} to enhance sensitivity to dynamic artifacts, while others explore the Mamba architecture~\cite{mamba, DeMamba} for its efficacy in long-sequence modeling to address temporal consistency. 
Furthermore, emerging research has begun to exploit the universal knowledge embedded in Large Multimodal Models, such as Qwen~\cite{qwen2.5, vidguard-r1}, attempting to construct universal detectors that identify logical fallacies in generated content through the model's comprehensive world understanding.

However, despite the significant strides made in semantic mining~\cite{busterx}, spatiotemporal sequence modeling~\cite{bai2024aigeneratedvideodetectionspatiotemporal}, and physical law verification~\cite{nsg-vd}, these approaches invariably overlook the critical role of frequency features. 
In contrast to high-level semantic anomalies, the low-level spectral artifacts introduced during the synthesis process are often far more resilient to mitigation~\cite{frank2020leveraging, luo2021generalizing}; yet, this pivotal dimension remains largely underexplored within general AI-generated video detection frameworks.

\textbf{Frequency-based Forensics.} 
Frequency analysis has long served as a robust instrument for distinguishing between real and synthesized imagery~\cite{lyu2005realistic, fridrich2012rich}. 
Recent studies in AI-generated image detection have further corroborated the potential of this dimension in uncovering generative traces~\cite{f3net, bihpf}. 
For instance, SPAI~\cite{spai} observes that real natural images adhere to specific spectral statistical patterns, whereas generated content often deviates from this distribution. 
Similarly, the Dual Frequency Branch framework~\cite{yan2025dualfrequencybranchframework} highlights that the phase spectrum encodes structural information that is often more discriminative than the amplitude spectrum alone. 
These findings suggest that spectral anomalies introduced by synthesis mechanisms are fundamental forensic cues.

However, transposing these image-based frequency priors to AI-generated video detection faces inherent barriers~\cite{dolhansky2020deepfake}. 
Unlike static images, video data is characterized by complex scene dynamics, such as object motion, lighting variations, and natural blur caused by camera movement. 
These inherent visual fluctuations introduce significant environmental noise into the frequency domain, which tends to submerge the already subtle and volatile generative artifacts~\cite{jiang2021focal}. 
The intense dynamic interference renders the direct extraction of reliable frequency features from video frames exceptionally difficult. 
Consequently, how to robustly isolate truly discriminative spectral clues amidst such dynamic video environments remains an unaddressed challenge in the field.

% \textbf{AI-Generated Video Detection Dataset.}
% Distinct from deepfakes, AIGVD targets fully synthetic content. 
% While several benchmarks have been proposed~\cite{wen2026busterxunifiedcrossmodalaigenerated, ji2024distinguishfakevideosunleashing, ni2025genvidbench6millionbenchmarkaigenerated, DeMamba}, they largely rely on earlier generators, failing to capture the high fidelity of recent commercial systems like Sora~\cite{openai2025sora2blog} and Kling~\cite{kuaishou2025kling}. 
% Our work addresses this by constructing a benchmark specifically targeting these state-of-the-art models to ensure up-to-date evaluation.

\section{Methodology}

% \begin{figure}[ht] % [t] 表示将图片置于当前页或下一页的顶部
%     \centering % 图片居中
%     % width=0.8\linewidth 控制图片宽度，可以根据实际视觉效果调整 0.7~0.9
%     % 如果你的 PDF 依然有白边，可以使用 trim 参数手动裁剪：
%     % trim=左 下 右 上，必须配合 clip 使用
%     \includegraphics[width=0.8\linewidth, clip]{framework.pdf} 
    
%     \caption{The overall architecture of SpecSem-Net.}
%     \label{fig:framework} % 用于文中引用，如 As shown in Fig. \ref{fig:framework}
% \end{figure}

\subsection{Overview}
The overall architecture of \textbf{SpecSem-Net} is illustrated in Figure \ref{fig:framework}. Unlike previous works that rely solely on semantic features, we explicitly incorporate spectral features to capture subtle generative artifacts~\cite{frank2020leveraging}. Meanwhile, our framework employs semantic features to filter out benign environmental content within these spectral features. We describe the detail of our proposed method as follows.

%Firstly, we intercept the final 2-second segment of a video and uniformly extract 8 frames as input. 
The video to detect is processed through two parallel branches. The \textbf{Spectral Module} first utilizes Fast Fourier Transform (FFT) and high-pass masking techniques to strip away dominant low-frequency semantic content, thereby explicitly exposing subtle, periodic frequency-domain artifacts. Simultaneously, the \textbf{Semantic Module} takes the raw frames directly as input and provides layer-by-layer semantic representations via a pre-trained CLIP encoder~\cite{clip}. Another CLIP encoder with the same architecture is applied to extract layer-by-layer spectral features. Further, to distinguish synthetic artifacts in spectral features from natural high-frequency textures (e.g., fine foliage or hair), we introduce a \textbf{Gated Merging Mechanism}. This mechanism leverages semantic guidance to adaptively modulate the spectral features~\cite{hu2018squeeze} layer by layer, ensuring that the model focuses on generative traces. Upon obtaining the fused frame-level features from the final layer, we employ a \textbf{Temporal Transformer~\cite{vaswani2017attention}} following the factorised encoder design~\cite{arnab2021vivit} to aggregate the temporal information of the frames. This module captures the evolution and continuity of artifacts along the temporal axis, ultimately producing the final detection prediction.

Formally, let $\mathcal{X} = \{x_i\}_{i=1}^T$ be a sequence of frames. 
For each frame $x_i$, the spectral branch and semantic branch extract initial features $F_{\text{spec}, i}^{(1)}$ and $F_{\text{sem}, i}^{(1)}$, respectively. 
These features undergo $L$ layers of deep interaction. 
At layer $l$, features are updated by $\text{CLIPBlock}_l(\cdot)$ before fusion. Let $\star \in \{\text{spec}, \text{sem}\}$:
\begin{align}
    \hat{F}_{\star, i}^{(l)} &= \text{CLIPBlock}_l \left( F_{\star, i}^{(l)} \right) \\
    F_{\text{spec}, i}^{(l+1)} &= \mathcal{G} \left( \hat{F}_{\text{spec}, i}^{(l)}, \hat{F}_{\text{sem}, i}^{(l)} \right)
    \label{eq:gated_merging}
\end{align}
The final refined frame-level representation is denoted as $h_i = \text{Pool}(F_{\text{spec}, i}^{(L+1)})$, where $\text{Pool}(\cdot)$ transforms the spatial feature map into a compact vector. The video-level prediction $\hat{y}$ is then formulated as:
\begin{equation}
    \hat{y} = \sigma \left( \text{FC} \left( \mathcal{T} (h_1, \dots, h_T) \right) \right)
    \label{eq:overall_prediction}
\end{equation}
where $\mathcal{T}(\cdot)$ is the Temporal Transformer~\cite{vaswani2017attention} that aggregates the sequence $\{h_i\}_{i=1}^T$ to produce the final probability of the video being AI-generated.

% \begin{figure}[h]
%     \centering
%     \includegraphics[width=0.9\linewidth]{Spectral Feature Extractoin.pdf}
%     \caption{Detailed pipeline of the Spectral Feature Extraction module.}
%     \label{fig:Spectral Feature Extraction}
% \end{figure}

\subsection{Spectral Input Construction}

To capture the frequency flaws left by generative models during the image reconstruction process, we design a spectral extraction module based on high-pass filtering. This module leverages frequency-domain filtering to suppress semantic information, thereby better focusing on high-frequency artifacts.

\subsubsection{Domain Transformation}

For each sampled frame $x_i \in \mathbb{R}^{3 \times H \times W}$, we first utilize the two-dimensional Fast Fourier Transform (2D FFT~\cite{cooley1965algorithm}) to convert it into the frequency domain:

\begin{equation}
X_i = \mathcal{F}(x_i)
\end{equation}

To facilitate subsequent filtering operations, we apply the \texttt{fftshift} operator to move the low-frequency components of the spectrum to the center of the spectral matrix. In the transformed spectrum $X_i$, the central region represents high-level semantics such as object contours and colors, while the peripheral regions contain fine textures and the periodic noise characteristic of generative models.

\subsubsection{High-Pass Mask Filtering}

Based on the prior knowledge that generative artifacts are typically hidden in high-frequency bands~\cite{dzanic2020fourier, chandrasegaran2021closer}, we construct a binary high-pass mask $M \in \{0, 1\}^{H \times W}$. The logic of the mask is as follows:

\begin{equation}
M(u, v) = \begin{cases} 0, & \text{if } \sqrt{(u-c_y)^2 + (v-c_x)^2} \leq r \\ 1, & \text{otherwise} \end{cases}
\end{equation}

where $(c_y, c_x)$ are the coordinates of the spectral center, and $r$ is the set filtering radius. By performing element-wise multiplication between the spectrum and the mask ($X'_i = X_i \odot M$), we filter out the dominant low-frequency components, forcing the model to focus on the obscured high-frequency residuals.

\subsubsection{Spectral Residual Reconstruction}

Finally, we transform the processed spectrum back to the spatial domain via the Inverse Fast Fourier Transform (iFFT) to obtain the spectral residual map $R_i$:

\begin{equation}
R_i = \text{Re}(\mathcal{F}^{-1}(\text{ifftshift}(X'_i)))
\end{equation}

The real part is taken here to suppress minute imaginary noise during computation. The reconstructed $R_i$  highlights grid artifacts~\cite{odena2016deconvolution} and texture anomalies within the video frames. Subsequently, $R_i$ is fed as input into the trainable spectral branch (Trainable CLIP~\cite{clip}) for deep feature extraction.

% \begin{figure}[h]
%     \centering
%     \includegraphics[width=1.1\linewidth]{figure/gate.pdf}
%     \caption{Detailed pipeline of the Gated-Merging module.\textcolor{red}{change i to l}}
%     \label{fig:Gated-Merging}
% \end{figure}

\subsection{Spectral and Semantic Feature Extraction and Interaction}
After obtaining the dual-stream inputs of semantic and spectrum, effectively fusing these diverse features is critical for detecting video forgeries. We propose a layer-wise gated merging module, which aims to leverage rich semantic contexts to dynamically modulate the representation of spectral features.

\subsubsection{Feature Encoding and Gate Computation}
To capture deep feature representations within the $l$-th layer, we first feed the input features into the CLIP Encoder Block to obtain the intermediate encoded representations, denoted as $\hat{F}^{(l)}$:
\begin{equation}
\begin{split}
\hat{F}_{sem}^{(l)} &= \text{CLIPBlock}_l(F_{sem}^{(l)}), \\
\hat{F}_{spec}^{(l)} &= \text{CLIPBlock}_l(F_{spec}^{(l)})
\end{split}
\label{eq:encoding}
\end{equation}
Since the intensity of artifacts produced by generative models is often closely related to the image content (e.g., artifacts in smooth regions are more perceptible than those in complex texture regions~\cite{chai2020makes}), we design a gating mechanism to capture this dependency. We leverage the encoded intermediate features to generate the cross-modal gate. Specifically, we concatenate $\hat{F}_{sem}^{(l)}$ and $\hat{F}_{spec}^{(l)}$ along the channel dimension, followed by a lightweight Multi-Layer Perceptron (MLP) and a Sigmoid activation function to produce the gated weight vector $g^{(l)}$:
\begin{equation}
g^{(l)} = \sigma(\text{MLP}([\hat{F}_{sem}^{(l)}; \hat{F}_{spec}^{(l)}]))
\label{eq:gate}
\end{equation}
where $[\cdot; \cdot]$ denotes the concatenation operation, and the dimension of $g^{(l)}$ is consistent with the feature dimension. This weight vector reflects the semantic branch's assessment of the importance of each channel in the current spectral features.

\subsubsection{Dual-Branch Feature Evolution}
With the computed gate, we adaptively modulate the spectral branch. For the semantic branch, we directly use the encoded features for the next layer to maintain stable contextual updates:
\begin{equation}
F_{sem}^{(l+1)} = \hat{F}_{sem}^{(l)}
\label{eq:sem_update}
\end{equation}
Simultaneously, for the spectral branch, we formulate the modulation term as $(1 + g^{(l)})$~\cite{perez2018film} to preserve fundamental frequency details while highlighting artifacts:
\begin{equation}
F_{spec}^{(l+1)} = \hat{F}_{spec}^{(l)} \odot (1 + g^{(l)})
\label{eq:spec_update}
\end{equation}
where $\odot$ represents element-wise multiplication. By retaining the original features via the identity term $1$~\cite{he2016deep}, the spectral branch can preserve intrinsic frequency information while adaptively enhancing potential anomaly regions guided by the semantic context. This one-way guidance ensures precise localization of high-level frequency traces.

\subsection{Temporal Aggregation}

Given that current video generative models often struggle to maintain long-term spatiotemporal consistency, frame-level detectors are prone to overlooking dynamic artifacts across frames, such as texture flickering. To capture the subtle temporal cues, we incorporate a Temporal Transformer module~\cite{vaswani2017attention} following the frame-level feature extraction.

Specifically, the sequence of fused frame features, denoted as $H = \{h_1, h_2, ..., h_T\}$, is fed into a Transformer Encoder. This module utilizes Multi-Head Self-Attention to model global temporal dependencies.

\section{Experiments}

% \begin{table}[ht!]
%     \caption{Dataset statistics.}
%     \label{tab:dataset-stats}
%     \begin{center}
%         \begin{small}
%             \begin{sc}
%                 \begin{tabular}{lcc}
%                     \toprule
%                     Category & Training & Test \\
%                     \midrule
%                     Real (Kinetics) & 18K & 1K \\
%                     HunyuanVideo & 6K & - \\
%                     DynamiCrafter & 6K & - \\
%                     Latte & 6K & - \\
%                     Hailuo & - & 200 \\
%                     Sora & - & 200 \\
%                     Veo & - & 200 \\
%                     Kling & - & 200 \\
%                     Wan & - & 200 \\
%                     \midrule
%                     Total & 36K & 2K \\
%                     \bottomrule
%                 \end{tabular}
%             \end{sc}
%         \end{small}
%     \end{center}
%     \vskip -0.1in
% \end{table}

\subsection{Datasets}
\label{sec:datasets}

\textbf{Existing Benchmarks and Challenges.} 
While several AI generated video detection benchmarks have been proposed such as GenVideo~\cite{DeMamba}, GenVidBench~\cite{ni2025genvidbench6millionbenchmarkaigenerated}, BusterX++~\cite{wen2026busterxunifiedcrossmodalaigenerated} and GenVidDet~\cite{ji2024distinguishfakevideosunleashing}, they largely rely on earlier generators, failing to capture the high fidelity of recent commercial systems like Sora~\cite{openai2025sora2blog} and Kling~\cite{kuaishou2025kling}. Our work addresses this by constructing a benchmark specifically targeting these state-of-the-art models to ensure up-to-date evaluation.

\textbf{Experimental Scope.} To comprehensively evaluate the robustness and generalization capability of SpecSem-Net, our experiments are conducted on both our proposed more challenging benchmark and existing widely-used public dataset GenVideo~\cite{DeMamba}.

% \textbf{Data Sources and Composition.} To verify the generalization capability across unseen generative models, we construct a large-scale, multi-source AI-generated video detection dataset including both training and test splits.
% Our \textbf{Training Set} consists of 36,000 video clips. The authentic samples are sourced from the Kinetics-400 database~\cite{kay2017kinetics} to encompass a diverse range of macro-semantic scenarios. For the synthesized samples, we utilized several mainstream open-source models, including HunyuanVideo~\cite{cao2025hunyuanimage}, DynamiCrafter~\cite{xing2024dynamicrafter}, and Latte~\cite{ma2024latte}.
% Regarding the \textbf{Test Set}, we generate synthetic videos using advanced commercial generative models that have no overlap with those involved in the training split, namely Sora~\cite{openai2025sora2blog}, Kling~\cite{kuaishou2025kling}, Hailuo~\cite{hailuo}, Veo~\cite{deepmind2025veo}, and Wan~\cite{wan}. Detailed statistical distributions are provided in Table \ref{tab:dataset-stats}.

\textbf{Data Sources and Composition.} To verify the generalization capability across unseen generative models, we construct a large-scale, multi-source AI-generated video detection dataset. Specifically, our \textbf{Training Set} consists of 36,000 video clips, comprising 18,000 authentic samples from the Kinetics-400 database~\cite{kay2017kinetics} and 6,000 synthesized samples from each of the three mainstream open-source models (HunyuanVideo~\cite{cao2025hunyuanimage}, DynamiCrafter~\cite{xing2024dynamicrafter}, and Latte~\cite{ma2024latte}). For the \textbf{Test Set}, to rigorously evaluate generalization, we utilize 2,000 entirely unseen videos. This includes 1,000 authentic videos and 200 synthetic videos generated by each of the five advanced commercial models (Sora~\cite{openai2025sora2blog}, Kling~\cite{kuaishou2025kling}, Hailuo~\cite{hailuo}, Veo~\cite{deepmind2025veo}, and Wan~\cite{wan}).

\textbf{Generation Strategy.} We employ two methods to construct our synthetic data. First, to prevent the model from relying on ``content bias''---where the detector distinguishes real from fake based solely on macro-semantic inconsistencies---we developed a specific pipeline. In this pipeline, we uniformly sample authentic videos from 100 diverse action categories in Kinetics-400. Then, we utilize Gemini-2.5 Flash~\cite{team2023gemini} to extract fine-grained semantic descriptions from these authentic videos. These descriptions subsequently serve as text prompts to guide generators in producing video clips with highly similar content. The complete prompt templates are provided in Appendix \ref{sec:app_dataset}. Notably, our entire training set is generated using this pipeline. Second, we directly collect officially released demo videos from various model providers, since they represent generally higher quality samples that are worth showing. For the test set, 50\% of the samples are generated using the aforementioned pipeline, while the remaining 50\% consist of the collected demo videos. This hybrid construction ensures both semantic fairness and coverage of current state-of-the-art visual quality.

\subsection{Experimental Setup}

\textbf{Implementation Details.} All video frames are standardized to $224 \times 224$ pixels to ensure compatibility with pre-trained backbones. We utilize the ViT-B/32 variant of CLIP~\cite{clip} as a backbone for both semantic and spectral branches. For our temporal transformer block, to ensure sufficient representational capacity while maintaining computational efficiency, we instantiate the encoder with 8 attention heads and expand the feed-forward network dimension to four times the hidden dimension. Regarding the high-pass mask in the Spectral Module, we set the filtering radius to $r=32$ by default, which masks approximately 14.3\% of the low-frequency components. This empirical setting is theoretically supported to balance artifact retention and semantic suppression, with detailed sensitivity analysis provided in Appendix \ref{sec:app_ablations}. For model optimization, we employ the Adam optimizer~\cite{kingma2014adam} with a fixed learning rate of $1 \times 10^{-5}$ and a weight decay of $1 \times 10^{-4}$. The training process is conducted with a total batch size of 16, and we train the model for 15 epochs until convergence.

\textbf{Baselines.} To rigorously evaluate the effectiveness of SpecSem-Net, we compare it against a diverse set of state-of-the-art detectors: (1) \textbf{ResNet50}~\cite{he2016deep}, a classic CNN-based spatial artifact detector; (2) \textbf{VideoMAE}~\cite{videomae}, a Transformer-based model optimized for spatiotemporal representation; (3) \textbf{DeMamba}~\cite{DeMamba}, representing the latest Mamba-based architectures for AI-generated video detection; (4) \textbf{D3}~\cite{zheng2025d3}, a training-free AI-generated video detection model using second-order features; (5) \textbf{Qwen-2.5-VL}~\cite{qwen2.5}, a cutting-edge Multimodal Large Language Model (MLLM) and (6) \textbf{FreqNet}~\cite{freqnet}, a classic frequency-aware detector that learns forensic clues in the frequency space. We adapt it for video by averaging frame-level predictions. For a fair comparison, \textbf{all baseline models are re-trained on our training set using their respective official implementations, while D3~\cite{zheng2025d3} and Qwen-2.5-VL~\cite{qwen2.5} are evaluated directly without fine-tuning.}

\textbf{Evaluation Metrics.} Following the standard protocols in AI forensics, we report two key metrics: Accuracy (Acc) and F1-score (F1).

% \begin{figure*}[t] % 使用 figure* 可以跨双栏，如果图片较宽建议用这个
%   \centering
%   \includegraphics[width=\linewidth]{figure/figure3.pdf}
%   \caption{\textbf{Visualization of the Spark Case.} We visualize the feature evolution to demonstrate robustness against environmental noise. \textbf{(Col 1-2)} The high-pass filter inherently captures benign sparks as dominant high-frequency signals (\textit{Spectral Input}). \textbf{(Col 3)} Consequently, the \textit{Before Gate} features are heavily distracted by this noise. \textbf{(Col 4)} The \textit{Gate Mechanism}, guided by semantic context, identifies and down-weights these benign texture regions. \textbf{(Col 5)} As a result, the \textit{After Gate} features successfully suppress the spark interference, focusing on authentic generative artifacts.}
%   \label{fig:gate_vis}
% \end{figure*}

\begin{table*}[ht!] % 实验一
\centering
\setlength{\tabcolsep}{13.5pt}
\caption{Performance comparison on unseen frontier commercial generators. We evaluate SpecSem-Net against state-of-the-art detectors on videos generated by Wan, Kling, Veo, Sora, and Hailuo. \textbf{Bold} and \underline{underline} indicate the best and second-best results, respectively.}
\label{tab:exp1}
% 2. 修改列格式：Metric左右加竖线 (|c|)，Mean左边加竖线 (|c)
\resizebox{\linewidth}{!}{
\begin{tabular}{c c ccccc c}
\toprule
Method & Metric & Wan & Kling & Veo & Sora & Hailuo & Mean \\
\midrule
\multirow{3}{*}{ResNet50} 
    & Acc & \underline{80.48} & 78.94 & 81.98 & \underline{76.23} & \underline{80.23} & \underline{79.45} \\
    & F1  & \underline{82.80}  & 81.24 & 84.33 & \textbf{78.25} & \underline{82.55} & \underline{81.63} \\
    & AP  & \underline{81.82} & 89.66 & 69.19 & \underline{98.96} & \underline{86.99} & \underline{86.97} \\ 
\midrule
\multirow{3}{*}{DeMamba} 
    & Acc & 55.28 & 52.78 & 69.78 & 51.78 & 58.03 & 60.99 \\
    & F1  & 59.97 & 56.77 & 76.06 & 55.44 & 63.34 & 57.42 \\
    & AP  & 75.38 & 74.72 & 75.61 & 86.08 & 72.48 & 76.94 \\
\midrule
\multirow{3}{*}{D3} 
    & Acc & 57.81 & 56.81 & 58.06 & 52.31 & 54.56 & 55.91 \\
    & F1  & 70.01 & 69.08 & 70.24 & 64.72 & 66.94 & 68.23 \\
    & AP  & 60.59 & 76.95 & 48.60 & 72.05 & 67.52 & 66.00 \\
\midrule
\multirow{3}{*}{VideoMae} 
    & Acc & 76.16 & \underline{83.33} & \underline{84.41} & 67.91 & 74.66 & 78.11 \\
    & F1  & 74.73 & \underline{83.58} & \underline{89.12} & 62.72 & 72.70  & 77.09 \\
    & AP  & 75.53 & \underline{92.57} & 61.34 & 94.37 & 85.22 & 83.63 \\
\midrule
\multirow{3}{*}{Qwen-2.5} 
    & Acc & 60.76 & 62.76 & 50.26 & 60.23 & 59.76 & 59.06 \\
    & F1  & 41.64 & 46.21 & 12.34 & 40.45 & 39.25 & 36.69 \\
    & AP  & -     & -     & -     & -     & -     & -     \\
\midrule
\multirow{3}{*}{FreqNet} 
 & Acc & 65.76 & 60.35 & 68.98 & 61.14 & 58.91 & 63.03 \\
 & F1  & 72.94 & 68.14 & 75.25 & 68.66 & 67.45 & 70.50 \\
 & AP  & 75.26 & 74.36 & \underline{82.89} & 63.50 & 72.21 & 73.64 \\
\midrule
\multirow{3}{*}{Ours} 
    & Acc & \textbf{85.55} & \textbf{92.05} & \textbf{93.55} & \textbf{79.30} & \textbf{91.55} & \textbf{87.25}$_{\pm 0.96}$ \\
    & F1  & \textbf{85.09} & \textbf{92.31} & \textbf{93.85} & \underline{77.17} & \textbf{85.39} & \textbf{86.96}$_{\pm 0.76}$ \\
    & AP  & \textbf{93.63} & \textbf{96.50} & \textbf{81.59} & \textbf{99.64} & \textbf{93.65} & \textbf{95.01}$_{\pm 0.67}$ \\ 
\bottomrule
\end{tabular}
}
\end{table*} % 实验一

\begin{table*}[ht!]
\caption{Generalization performance on public benchmarks. The models are evaluated on seven mainstream open-source generation datasets to verify robustness. \textbf{Bold} and \underline{underline} indicate the best and second-best results, respectively.}
\centering
\label{tab:exp2}
\renewcommand{\arraystretch}{1.2}
% 【⚠️排版高危提醒】NeurIPS 官方模板强烈建议使用 booktabs 且“严禁包含竖线”。
% 如果提交系统报错，请将下方的 {c |c| ccccccc |c} 改为 {c c ccccccc c}
\resizebox{\linewidth}{!}{
\begin{tabular}{c c ccccccc c}
\toprule
Method & Metric & I2VGEN & \makecell{OPEN\\SORA} & PIKA & SEINE & SVD & \makecell{VIDEO\\CRAFTER} & \makecell{ZERO\\SCOPE} & Mean \\
\midrule
\multirow{3}{*}{ResNet50} 
 & Acc & 95.12 & 75.73 & 90.28 & 91.49 & 94.76 & \underline{93.05} & 68.60 & 87.08 \\
 & F1  & 93.76 & 58.77 & 86.76 & 88.60 & 93.27 & \underline{90.88} & 39.30 & 85.77 \\
 & AP  & \underline{99.88} & 95.34 & \underline{99.35} & \underline{99.74} & \underline{99.75} & \underline{99.68} & \textbf{98.11} & \underline{98.95} \\
\midrule
\multirow{3}{*}{DeMamba} 
 & Acc & \underline{95.16}& \underline{89.07}& \underline{92.51}& 94.10 & \underline{94.94} & 92.00 & \textbf{91.51} & \underline{92.76} \\
 & F1  & \underline{94.43}& \underline{86.46}& \underline{91.11}& 93.12 & \underline{94.16} & 90.44 & \textbf{89.80} & \underline{91.37} \\
 & AP  & 99.42 & \underline{96.07} & 98.27 & 98.89 & 99.45 & 97.34 & 96.07 & 98.09 \\
\midrule
\multirow{3}{*}{D3} 
 & Acc & 43.50 & 60.60 & 61.55 & 59.71 & 49.75 & 58.28 & 61.06 & 56.35 \\
 & F1  & 44.71 & 66.96 & 68.02 & 65.97 & 53.66 & 64.33 & 67.48 & 62.05 \\
 & AP  & 49.63 & 81.37 & 87.57 & 80.74 & 48.89 & 63.40 & 77.65 & 71.97 \\
\midrule
\multirow{3}{*}{VideoMae} 
 & Acc & 77.97 & 73.73 & 84.27 & \underline{94.61} & 93.34 & 71.98 & 84.56 & 83.07 \\
 & F1  & 65.12 & 55.43 & 77.36 & \underline{93.24} & 91.52 & 50.99 & 77.87 & 81.10 \\
 & AP  & 91.91 & 89.34 & 95.57 & 98.43 & 98.36 & 87.69 & 95.84 & 94.44 \\
\midrule
\multirow{3}{*}{Qwen-2.5} 
 & Acc & 76.39 & 62.50 & 65.48 & 73.30 & 73.40 & 58.05 & 86.46 & 70.80 \\
 & F1  & 65.50 & 31.26 & 40.02 & 59.15 & 59.37 & 16.27 & 82.76 & 53.54 \\
 & AP  & -     & -     & -     & -     & -     & -     & -     & -     \\
\midrule
\multirow{3}{*}{FreqNet} 
 & Acc & 72.27 & 68.16 & 71.02 & 72.71 & 72.72 & 73.02 & 65.52 & 70.77 \\
 & F1  & 77.73 & 73.55 & 76.48 & 78.16 & 78.17 & 78.46 & 70.71 & 76.18 \\
 & AP  & 85.31 & 76.33 & 84.97 & 85.02 & 87.68 & 89.83 & 78.10 & 83.89 \\
\midrule
\multirow{3}{*}{Ours} 
 & Acc & \textbf{97.40}& \textbf{94.08}& \textbf{96.96}& \textbf{96.49}& \textbf{97.42}& \textbf{97.32}& \underline{89.48}& \textbf{95.59}\\
 & F1  & \textbf{96.93}& \textbf{92.74}& \textbf{96.40}& \textbf{95.82}& \textbf{96.95}& \textbf{96.84}& \underline{86.32}& \textbf{95.45}\\
 & AP  & \textbf{99.95}& \textbf{98.63}& \textbf{99.76}& \textbf{99.79}& \textbf{99.94}& \textbf{99.87}& \underline{96.95}& \textbf{99.32}\\
\bottomrule
\end{tabular}
}
\end{table*}

\subsection{Results and Analysis}

\textbf{Performance on Our Constructed Dataset.} 
Table \ref{tab:exp1} presents the detection performance of various detectors on five unseen commercial generative models in our dataset. SpecSem-Net demonstrates great superiority, achieving a Mean Accuracy of 87.25\% and an F1 score of 86.96\%. Compared to the second-best method (ResNet50), we improve the accuracy by approximately 7.8\%.

It is worth analyzing the performance differences of different architectures when facing high-fidelity videos:

\paragraph{Collapse of Existing SOTA:} Models relying on strong spatiotemporal semantics, such as DeMamba~\cite{DeMamba} and VideoMAE~\cite{videomae}, suffer severe performance degradation when facing top-tier generators like Veo, Sora, and Kling. For instance, DeMamba's accuracy is only 69.78\% on Veo and merely 51.78\% on Sora (close to random guessing). This indicates that when frontier generative models can synthesize videos that are nearly perfect in macro-semantics and spatiotemporal coherence, detectors relying solely on semantic features become ineffective.

Interestingly, the relatively simple ResNet50 surpasses these complex architectures (Mean Accuracy 79.45\%). Since CNNs possess a stronger inductive bias towards local textures~\cite{geirhos2018imagenet}, we believe this characteristic enables ResNet50 to capture low-level spatial artifacts more effectively than semantics-focused models in this scenario.

\paragraph{Superiority of SpecSem-Net} In the highly challenging Veo and Kling data, SpecSem-Net maintains extremely high detection precision. This proves that by introducing gated frequency features, the model successfully spotted microscopic spectral fingerprints hidden beneath perfect visuals.

\textbf{Generalization on Public Benchmarks.} 
To further verify the universality and robustness of SpecSem-Net, we conducted evaluations on public benchmarks~\cite{DeMamba} containing seven mainstream generators, with results shown in Table \ref{tab:exp2}. Our method still achieves the best performance in the vast majority of test subsets, with a Mean Accuracy as high as 95.59\%.

Specifically, SpecSem-Net ranks first in 6 out of the 7 sub-datasets. Particularly on SVD (97.42\%), I2VGen (97.40\%), and VideoCrafter (97.32\%), the model achieves near-perfect detection. In contrast, other baseline methods exhibit some instability. VideoMAE performs mediocrely across multiple datasets. The consistent high performance of SpecSem-Net across datasets indicating the superiority of out proposed model.

\paragraph{Robustness and Efficiency.} Beyond clean benchmarks, AI-generated videos in the wild often suffer from severe degradation (e.g., social media compression, blurring). We evaluate the robustness of SpecSem-Net against varying levels of H.264/H.265 compression and Gaussian blur, demonstrating its superior stability over state-of-the-art baselines. Furthermore, our dual-stream architecture maintains highly competitive computational efficiency for practical deployment. Detailed quantitative results on robustness, inference latency (FPS), and computational complexity (FLOPs) are provided in Appendix \ref{sec:app_robustness} and \ref{sec:app_efficiency}.

\begin{table}[ht!] % 消融实验
    \caption{\textbf{Ablation Study.} We evaluate the contribution of each component on the proposed commercial benchmark. ``Sem.'' and ``Spec.'' denote the Semantic and Spectral branches, respectively. ``Gate'' denotes the Gated Merging Mechanism. \textbf{Bold} indicates the best performance.}
    \label{tab:ablation}
    \begin{center}
        \begin{small}
            \begin{sc}
                % 调整列间距
                \setlength{\tabcolsep}{5pt}
                \begin{tabular}{ccc|cc|cc}
                    \toprule
                    % 组件列：用打勾表示，清晰明了
                    \multicolumn{3}{c|}{Components} & \multicolumn{2}{c|}{Sora (Hardest)} & \multicolumn{2}{c}{Average} \\
                    Sem. & Spec. & Gate & Acc & F1 & Acc & F1 \\
                    \midrule
                    % 1. Semantic Only
                    \checkmark & & & 52.45& 59.38& 55.68& 63.05\\
                    % 2. Spectral Only
                    & \checkmark & & 63.12& 71.49& 70.34& 62.04\\
                    % 3. Without Gate (双流直接拼接，没有 Gate)
                    \checkmark & \checkmark & & 70.76& 74.18& 74.54& 78.14\\
                    \midrule
                    % 4. Ours (全都有)
                    \checkmark & \checkmark & \checkmark & \textbf{79.30} & \textbf{77.17} & \textbf{87.25} & \textbf{86.96} \\
                    \bottomrule
                \end{tabular}
            \end{sc}
        \end{small}
    \end{center}
    \vskip -0.1in
\end{table}

\subsection{Ablation Study}

To verify the effectiveness of the core components within SpecSem-Net, we conducted a detailed ablation study on our constructed benchmark. Table \ref{tab:ablation} presents the quantitative results under different configurations, including the most challenging \textit{Sora} subset and the overall average metrics.

\paragraph{Complementarity of Semantic and Spectral Branches} First, we evaluate the performance of the two single-stream branches independently. As can be seen, 
model relying solely on semantic features (\textit{Semantic Only}) performs the worst, yielding an average accuracy of only 55.68\%. Crucially, on the Sora dataset, its accuracy drops to 52.45\%, which is nearly equivalent to random guessing. This strongly corroborates our hypothesis that for high-fidelity videos with perfect semantics, traditional semantic detectors are rendered completely ineffective~\cite{liu2024turnsimrealrobust}.
In contrast, the model relying solely on spectral features (\textit{Spectral Only}) exhibits better performance, with the average accuracy improving to 70.34\%. This suggests that despite the deceptive visual content, discriminative artifacts persist in the frequency domain.

\textbf{Critical Role of Gated Merging Mechanism.} When we directly concatenate the two branches (\textit{w/o Gate}), the performance improves to 74.54\%. Although this outperforms single-stream models, the margin of improvement is limited. This is because natural videos inherently contain abundant high-frequency information (e.g., hair, foliage)~\cite{chandrasegaran2021closer}. Simple concatenation fails to distinguish between ``natural high-frequency textures'' and ``generated artifacts'', leading to potential interference for detection. 
Upon introducing the Gated Merging Mechanism (\textit{Ours}), the model's performance achieves a giant leap. The average accuracy surges from 74.54\% to 87.25\% (an improvement of approximately 12.7\%), and the accuracy on Sora also increases by nearly 9\%. This demonstrates that the gating mechanism effectively leverages semantic context to ``filter'' spectral noise, enabling the model to focus on authentic forgery traces.

\paragraph{Architectural and Hyperparameter Choices.} To further justify our architectural design, we conduct additional ablation studies on the choice of the temporal aggregation head (comparing Mean Pooling, 3D-CNN, and Temporal Transformer) and the selection of the high-pass filtering radius $r$. The detailed results and theoretical justifications are presented in Appendix \ref{sec:app_ablations}.

% \subsection{Qualitative Analysis} 
% To intuitively validate the robustness of SpecSem-Net against environmental noise, we visualize a representative sample featuring intense high-frequency texture interference (i.e., splashing sparks) in Figure \ref{fig:gate_vis}. 
% As illustrated in the second column (\textit{Spectral Input}), while the fixed high-pass filter is effective at extracting high-frequency components, it inevitably retains sharp environmental edges, such as the contours of the sparks. 
% This observation explains why naive frequency analysis often fails: \textit{authentic physical textures} and \textit{generative artifacts} are typically entangled within the high-frequency domain. 
% Observing the third column (\textit{Before Gate}), it is evident that without semantic guidance, the intermediate features are heavily distracted by this noise, with activations predominantly concentrated on the benign spark regions. 
% However, upon introducing the Gated Merging Mechanism (\textit{After Gate}), activations in the spark regions are significantly suppressed. 
% This demonstrates that the Gate module functions as a \textbf{``Semantic Filter''}: it leverages contextual information from the semantic branch to identify these high-frequency signals as natural physical phenomena rather than forgery traces, thereby achieving precise feature denoising.

\paragraph{Qualitative Insights.} To intuitively demonstrate how our Gated Merging Mechanism acts as a semantic filter against intense high-frequency environmental noise (e.g., splashing sparks), we provide detailed qualitative visualizations and feature evolution analysis in Appendix \ref{sec:app_qualitative}.

% \subsection{Investigation of Trace Accumulation} \textcolor{red}{need to rerun}

% \begin{table}[ht!] % 采样消融
%     \caption{\textbf{Impact of Sampling Strategy. \textcolor{red}{need to rerun}} 
%     We compare different frame sampling strategies on detection performance. 
%     Our strategy (Tail 30\%) captures the most discriminative artifacts. 
%     $\Delta$ denotes the improvement over the baseline (Front 30\%).}
%     \label{tab:sampling_strategy}
%     \begin{center}
%         \begin{small}
%             \begin{sc} % Small Caps 字体
%                 % 调整列间距，如果表格太窄可以加大这个数字
%                 \setlength{\tabcolsep}{5pt} 
                
%                 \begin{tabular}{lcc}
%                     \toprule
%                     Sampling Strategy & Acc (\%) & F1 (\%) \\
%                     \midrule
%                     Front 2s Frames & 80.60 & 80.53 \\
%                     Random Sampling   & 83.62 & 82.05 \\
%                     \midrule
%                     % 使用 \textbf 加粗“Ours”这一行，强调你的方法
%                     \textbf{Tail 2s Frames (Ours)} & \textbf{86.6} & \textbf{86.6} \\
%                     \midrule
%                     % Delta 行，通常用灰色或者普通字体表示增益
%                     % 这里的 \Delta 需要在数学模式下
%                     $\Delta$ (Improvement) & +6 & +5.07 \\
%                     \bottomrule
%                 \end{tabular}
%             \end{sc}
%         \end{small}
%     \end{center}
%     \vskip -0.1in % 减少表格到底部的空白
% \end{table}

\section{Conclusion and Limitations}

In this paper, we propose \textbf{SpecSem-Net} for effective AI-generated video detection. SpecSem-Net  addresses the critical limitation of existing detectors that often fail on high-fidelity videos generated by powerful models such as Sora due to an over-reliance on semantic features. By introducing spectral features and further leveraging a Gated Merging Mechanism to apply semantic features to adaptively modulate extracted spectral features, SpecSem-Net is capable of precisely capturing  forgery traces within the spectral domain. We then construct a comprehensive benchmark comprising videos generated by five frontier commercial generators to rigorously evaluate the generalization capability of AI-generated video detectors. Experimental results demonstrate that SpecSem-Net achieves an average accuracy of 95.59\% on existing widely-used benchmark and 87.25\% on our proposed more challenging benchmark, surpassing all existing state-of-the-art methods. Ablation studies further prove the effectiveness of introducing both spectral and semantic features and the usefulness of gated merging mechanism. 

\paragraph{Limitations and Future Work.} 
Despite the gains, SpecSem-Net currently performs video-level predictions and does not yet support pixel-level localization of artifacts. Additionally, the fixed filtering radius ($r=32$) could be further optimized through adaptive frequency masking in future work. While robust to common degradations, extreme blurring can still impact high-frequency feature extraction, which we aim to address in the future.

\small
\bibliographystyle{unsrt} % unsrt 表示按照正文引用顺序排版
\bibliography{example_paper}

%%%%%%%%%%%%%%%%%%%%%%%%%%%%%%%%%%%%%%%%%%%%%%%%%%%%%%%%%%%%

\appendix

% ---------------- 模块二：鲁棒性测试 ----------------
\section{Robustness in Real-World Scenarios}
\label{sec:app_robustness}

As raised in previous discussions, high-frequency spectral artifacts are inherently susceptible to severe video compression or blurring. To evaluate our model's operational boundaries in real-world scenarios, we conducted comprehensive robustness tests. We applied varying levels of H.264 compression (CRF 18, 23), H.265 compression (CRF 18, 23), and Gaussian blur (Sigma 0.5, 1.0) to our test set. 

As shown in Table \ref{tab:robustness}, under common compressions (CRF 18/23), SpecSem-Net significantly outperforms all baselines across all metrics. Even under severe blurring (Sigma 1.0) which heavily destroys fine-grained spectral artifacts, our model maintains the highest Average Precision (AP), indicating a highly stable overall predictive performance compared to models that overly rely on spatial semantics.

\begin{table}[ht!]
    \caption{Robustness evaluation under real-world video degradations on our commercial benchmark. \textbf{Bold} and \underline{underline} indicate the best and second-best results, respectively.}
    \label{tab:robustness}
    \centering
    \small
    \renewcommand{\arraystretch}{1.1}  
    \begin{tabular}{ll cccc cc}
        \toprule
        \multirow{2}{*}{Method} & \multirow{2}{*}{Metric} & \multicolumn{2}{c}{H.264} & \multicolumn{2}{c}{H.265} & \multicolumn{2}{c}{Gaussian Blur} \\
        \cmidrule(lr){3-4} \cmidrule(lr){5-6} \cmidrule(lr){7-8}
        & & CRF 18 & CRF 23 & CRF 18 & CRF 23 & Sig 0.5 & Sig 1.0 \\
        \midrule
        \multirow{3}{*}{ResNet50} 
        & Acc & \underline{77.42} & \underline{74.49} & \underline{76.77} & \underline{74.24} & \textbf{79.75} & \textbf{73.65} \\
        & F1  & \underline{77.24} & 72.19 & \underline{77.52} & \underline{72.58} & \underline{80.70} & \textbf{77.06} \\
        & AP  & \underline{85.27} & \underline{84.46} & \underline{83.93} & 82.92 & \underline{86.76} & \underline{84.57} \\
        \midrule
        \multirow{3}{*}{DeMamba} 
        & Acc & 60.27 & 60.94 & 58.07 & 57.78 & 55.68 & 55.58 \\
        & F1  & 57.68 & 58.06 & 56.72 & 56.42 & 54.63 & 54.61 \\
        & AP  & 78.72 & 78.09 & 75.26 & 75.09 & 78.93 & 78.28 \\
        \midrule
        \multirow{3}{*}{D3} 
        & Acc & 63.62 & 62.98 & 62.03 & 62.18 & 63.57 & 63.18 \\
        & F1  & 69.55 & 69.55 & 69.14 & 69.00 & 69.57 & 69.26 \\
        & AP  & 69.28 & 69.12 & 69.32 & 69.77 & 68.95 & 68.88 \\
        \midrule
        \multirow{3}{*}{VideoMAE} 
        & Acc & 69.83 & 68.93 & 69.78 & 69.28 & 70.37 & \underline{70.62} \\
        & F1  & 58.92 & 56.83 & 59.32 & 57.75 & 60.59 & 61.31 \\
        & AP  & 84.38 & 83.95 & 83.64 & \underline{83.36} & 84.32 & 83.86 \\
        \midrule
        \multirow{3}{*}{Qwen-2.5} 
        & Acc & 56.91 & 56.76 & 57.90 & 57.60 & 57.50 & 57.35 \\
        & F1  & 37.76 & 37.41 & 40.19 & 39.44 & 39.21 & 38.87 \\
        & AP  & -     & -     & -     & -     & -     & -     \\
        \midrule
        \multirow{3}{*}{FreqNet}
        & Acc & 65.71 & 64.96 & 64.86 & 63.97 & 62.93 & 62.48 \\
        & F1  & 72.68 & \underline{72.27} & 71.88 & 71.51 & 71.39 & 71.66 \\
        & AP  & 76.50 & 74.07 & 75.92 & 74.46 & 75.14 & 77.49 \\
        \midrule
        \multirow{3}{*}{Ours} 
        & Acc & \textbf{83.77} & \textbf{82.03} & \textbf{79.70} & \textbf{78.76} & \underline{79.31} & 64.12 \\
        & F1  & \textbf{84.26} & \textbf{81.47} & \textbf{81.38} & \textbf{78.75} & \textbf{82.28} & \underline{73.05} \\
        & AP  & \textbf{92.59} & \textbf{90.65} & \textbf{90.11} & \textbf{87.33} & \textbf{93.79} & \textbf{91.08} \\
        \bottomrule
    \end{tabular}
\end{table}

% ---------------- 模块二：计算开销 ----------------
\section{Computational Complexity and Inference Latency}
\label{sec:app_efficiency}

To demonstrate the feasibility of SpecSem-Net for real-time or high-throughput deployment, we benchmarked the inference latency (FPS) and computational complexity (FLOPs) against baseline methods. All evaluations were conducted on a single NVIDIA RTX 3090 GPU. As detailed in Table \ref{tab:efficiency}, the computational burden of our dual-stream architecture (74.23G FLOPs) is highly competitive, outperforming complex spatiotemporal architectures like DeMamba and D3, while achieving the highest detection accuracy.

\begin{table}[ht!]
    \caption{Computational complexity and inference latency (measured on RTX 3090).}
    \label{tab:efficiency}
    \centering
    \begin{tabular}{lccc}
        \toprule
        Method & FLOPs (G) $\downarrow$ & FPS $\uparrow$ & Acc (\%) $\uparrow$ \\
        \midrule
        ResNet50      & 32.88  & \textbf{1127.91} & 79.45 \\
        DeMamba       & 147.48 & 228.95  & 60.99 \\
        D3            & 282.45 & 361.10  & 55.91 \\
        VideoMAE      & 135.18 & 483.36  & 78.11 \\
        FreqNet       & \textbf{15.79} & 246.94 & 63.03 \\
        \midrule
        Ours & 74.23  & 475.92  & \textbf{87.25} \\
        \bottomrule
    \end{tabular}
\end{table}

% ---------------- 模块三：额外参数消融 ----------------
\section{Additional Ablation Studies}
\label{sec:app_ablations}

\subsection{Temporal Aggregation Head}
To justify our choice of the Temporal Transformer for frame-level feature aggregation, we compared it against Mean Pooling and a 3D-CNN on our commercial benchmark. As shown in Table \ref{tab:temporal_head}, while the Temporal Transformer yields the highest accuracy (87.25\%), the marginal performance gap among the three heads (all $>85\%$) leads to a critical conclusion: the temporal aggregation head is not the bottleneck. The fact that even a simple Mean Pooling achieves highly competitive accuracy demonstrates the inherent superiority of our dual-stream backbone and the Gated Merging Mechanism.

\begin{table}[ht!]
    \caption{Ablation on different temporal aggregation heads.}
    \label{tab:temporal_head}
    \centering
    \begin{tabular}{lc}
        \toprule
        Method & Accuracy (\%) \\
        \midrule
        Mean Pooling & 85.91 \\
        3D-CNN       & 86.75 \\
        Transformer  & \textbf{87.25} \\
        \bottomrule
    \end{tabular}
\end{table}

\subsection{High-pass Mask Radius}
Table \ref{tab:radius} presents the sensitivity analysis for the high-pass filtering radius $r$. For our $224 \times 224$ input resolution, we set the default radius to $r=32$, which equates to an approximately 14.3\% mask ratio. This setting achieves the optimal balance for Accuracy and F1-score and is theoretically supported by the findings in Doloriel et al. (2024), which concluded that a $\sim15\%$ frequency masking ratio yields peak forensic detection performance.

\begin{table}[ht!]
    \caption{Ablation on the high-pass mask radius $r$.}
    \label{tab:radius}
    \centering
    \begin{tabular}{ccccc}
        \toprule
        Radius ($r$) & Mask Ratio (\%) & Mean Acc (\%) & Mean F1 (\%) & Mean AP (\%) \\
        \midrule
        0 (All-pass) & 0\%      & 83.33 & 84.70 & 93.76 \\
        16           & $\sim$3.6\%  & 83.37 & 84.70 & 93.68 \\
        32 (Ours)    & $\sim$14.3\% & \textbf{87.25} & \textbf{86.96} & 93.94 \\
        112          & 50\%     & 86.75 & 86.29 & \textbf{94.50} \\
        \bottomrule
    \end{tabular}
\end{table}

\section{Detailed Dataset Generation}
\label{sec:app_dataset}

To ensure the semantic fairness of our constructed benchmark and strictly evaluate the detectors' ability to capture low-level generative artifacts rather than high-level visual anomalies (e.g., impossible physics or bizarre subjects), we developed an automated text-to-video generation pipeline.

\paragraph{Semantic Extraction Prompt.}
We first uniformly sample authentic videos from the Kinetics-400 database~\cite{kay2017kinetics}. For each video, we utilize the Gemini-2.5 Flash Multimodal Large Language Model (MLLM)~\cite{team2023gemini} to extract a fine-grained, objective text description. To standardize the output, we query the MLLM with the following unified prompt template:

\begin{quote}
    \textit{``Please describe this video in a single, highly detailed sentence. Focus strictly on three aspects: (1) the main subject's appearance and action, (2) the background environment, and (3) the camera motion (e.g., panning, zooming, or static). Do not include subjective evaluations.''}
\end{quote}

\paragraph{Video Generation Prompt.}
The text descriptions generated by Gemini (e.g., \textit{``A golden retriever is running happily across a lush green park while the camera slowly pans to the right''}) are then directly utilized as the input text prompts for both the open-source generators and the commercial generators. 

\section{Qualitative Analysis on Environmental Noise}
\label{sec:app_qualitative}

\begin{figure}[ht!]
  \centering
  % 如果图片在附录里显得太宽，可以将 1.0 改小一点，比如 0.9
  \includegraphics[width=1.0\linewidth]{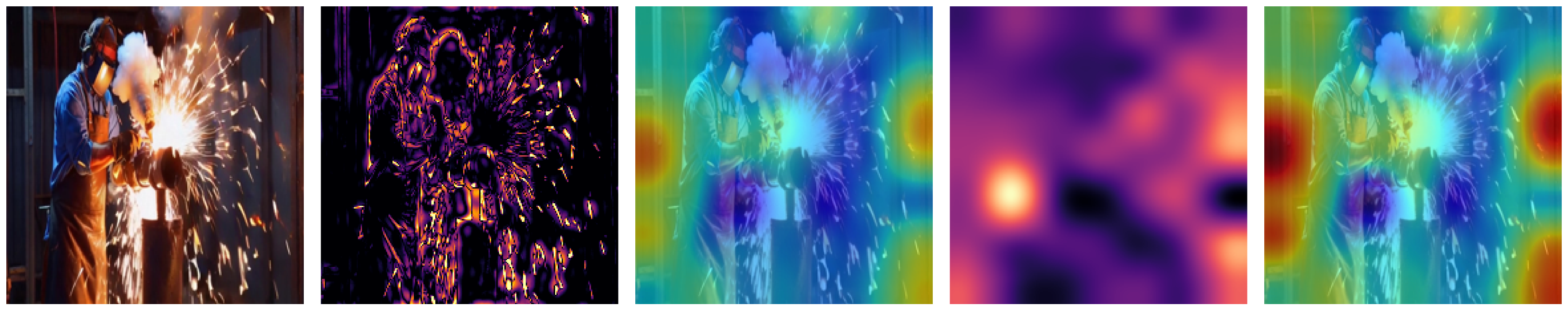}
  \caption{\textbf{Visualization of the Spark Case.} We visualize the feature evolution to demonstrate robustness against environmental noise. (Col 1-2) The high-pass filter inherently captures benign sparks as dominant high-frequency signals. (Col 3) Consequently, the features before the gating mechanism are heavily distracted by this noise. (Col 4) The Gated Merging Mechanism identifies and down-weights these benign texture regions. (Col 5) As a result, the suppressed features successfully focus on authentic generative artifacts.}
  \label{fig:gate_vis_app}
\end{figure}

To validate the robustness of SpecSem-Net against environmental noise, we visualize a representative sample featuring intense high-frequency texture interference (i.e., splashing sparks) in Figure \ref{fig:gate_vis_app}. 

As illustrated in Figure \ref{fig:gate_vis_app}, while the fixed high-pass filter is effective at extracting high-frequency components, it inevitably retains sharp environmental edges, such as the contours of the sparks. This observation explains why naive frequency analysis often fails: authentic physical textures and generative artifacts are typically entangled within the high-frequency domain. 

Observing the third column, it is evident that without semantic guidance, the intermediate features are heavily distracted by this noise, with activations predominantly concentrated on the benign spark regions. However, upon introducing the Gated Merging Mechanism (After Gate), activations in the spark regions are significantly suppressed. This demonstrates that the Gate module functions as a \textbf{``Semantic Filter''}: it leverages contextual information from the semantic branch to identify these high-frequency signals as natural physical phenomena rather than forgery traces, thereby achieving precise feature denoising.

\section{Broader Impact Statement}
\label{sec:app_impact}

This work mainly aims at detecting AI-generated videos to mitigate the risks of disinformation. Yet we acknowledge that no detector is perfect and our proposed detector may also make mistakes. So our proposed model is for research only and should not be used to definitively criticize whether a video is generated by AI. The generated videos contained in our benchmark are also for research only and do not aim at conveying any real-world information. We source real videos and demos from open-source resources and strictly follow their copyright requirements.

%%%%%%%%%%%%%%%%%%%%%%%%%%%%%%%%%%%%%%%%%%%%%%%%%%%%%%%%%%%%

\newpage
\section*{NeurIPS Paper Checklist}

\begin{enumerate}

\item {\bf Claims}
    \item[] Question: Do the main claims made in the abstract and introduction accurately reflect the paper's contributions and scope?
    \item[] Answer: \answerYes{} % Replace by \answerYes{}, \answerNo{}, or \answerNA{}.
    \item[] Justification: We clearly state our contributions, including the proposed SpecSem-Net, the newly constructed benchmark, and the corresponding experimental results.

\item {\bf Limitations}
    \item[] Question: Does the paper discuss the limitations of the work performed by the authors?
    \item[] Answer: \answerYes{} % Replace by \answerYes{}, \answerNo{}, or \answerNA{}.
    \item[] Justification: We explicitly discuss the limitations of our model, such as the fixed filtering radius and the vulnerability to severe blurring, in the "Conclusion and Limitations" section.

\item {\bf Theory assumptions and proofs}
    \item[] Question: For each theoretical result, does the paper provide the full set of assumptions and a complete (and correct) proof?
    \item[] Answer: \answerNA{} % Replace by \answerYes{}, \answerNo{}, or \answerNA{}.
    \item[] Justification: Our paper is an empirical study focusing on AI-generated video detection and does not include theoretical proofs.

    \item {\bf Experimental result reproducibility}
    \item[] Question: Does the paper fully disclose all the information needed to reproduce the main experimental results of the paper to the extent that it affects the main claims and/or conclusions of the paper (regardless of whether the code and data are provided or not)?
    \item[] Answer: \answerYes{} % Replace by \answerYes{}, \answerNo{}, or \answerNA{}.
    \item[] Justification: We detailed our experimental settings, including hyperparameters and dataset construction strategies, in Section 4.1, 4.2, and Appendix D.

\item {\bf Open access to data and code}
    \item[] Question: Does the paper provide open access to the data and code, with sufficient instructions to faithfully reproduce the main experimental results, as described in supplemental material?
    \item[] Answer: \answerYes{} % Replace by \answerYes{}, \answerNo{}, or \answerNA{}.
    \item[] Justification: As stated in the footnote of Section 1, our code and constructed benchmark will be made publicly available upon acceptance to facilitate future research.

\item {\bf Experimental setting/details}
    \item[] Question: Does the paper specify all the training and test details (e.g., data splits, hyperparameters, how they were chosen, type of optimizer) necessary to understand the results?
    \item[] Answer: \answerYes{} % Replace by \answerYes{}, \answerNo{}, or \answerNA{}.
    \item[] Justification: We provide comprehensive details on the datasets, evaluation metrics, and implementation specifics in Section 4.1, 4.2, and Appendix D.

\item {\bf Experiment statistical significance}
    \item[] Question: Does the paper report error bars suitably and correctly defined or other appropriate information about the statistical significance of the experiments?
    \item[] Answer: \answerYes{} % Replace by \answerYes{}, \answerNo{}, or \answerNA{}.
    \item[] Justification: We evaluated SpecSem-Net across 5 different random seeds and reported the mean and standard deviation ($\pm$) in Table 1 to ensure statistical stability.

\item {\bf Experiments compute resources}
    \item[] Question: For each experiment, does the paper provide sufficient information on the computer resources (type of compute workers, memory, time of execution) needed to reproduce the experiments?
    \item[] Answer: \answerYes{} % Replace by \answerYes{}, \answerNo{}, or \answerNA{}.
    \item[] Justification: We report the hardware details (a single NVIDIA RTX 3090 GPU), inference latency (FPS), and computational complexity (FLOPs) in Appendix B.
    
\item {\bf Code of ethics}
    \item[] Question: Does the research conducted in the paper conform, in every respect, with the NeurIPS Code of Ethics \url{https://neurips.cc/public/EthicsGuidelines}?
    \item[] Answer: \answerYes{} % Replace by \answerYes{}, \answerNo{}, or \answerNA{}.
    \item[] Justification: We have reviewed and strictly adhered to the NeurIPS Code of Ethics throughout our research and dataset construction.

\item {\bf Broader impacts}
    \item[] Question: Does the paper discuss both potential positive societal impacts and negative societal impacts of the work performed?
    \item[] Answer: \answerYes{} % Replace by \answerYes{}, \answerNo{}, or \answerNA{}.
    \item[] Justification: We discuss the positive impacts of mitigating disinformation and the potential risks of AI-generated videos in the "Broader Impact Statement" in Appendix F.
    
\item {\bf Safeguards}
    \item[] Question: Does the paper describe safeguards that have been put in place for responsible release of data or models that have a high risk for misuse (e.g., pre-trained language models, image generators, or scraped datasets)?
    \item[] Answer: \answerNA{} % Replace by \answerYes{}, \answerNo{}, or \answerNA{}.
    \item[] Justification: Our work focuses on a detection model rather than a generative model, and therefore does not introduce high-risk assets that require specific safeguards.

\item {\bf Licenses for existing assets}
    \item[] Question: Are the creators or original owners of assets (e.g., code, data, models), used in the paper, properly credited and are the license and terms of use explicitly mentioned and properly respected?
    \item[] Answer: \answerYes{} % Replace by \answerYes{}, \answerNo{}, or \answerNA{}.
    \item[] Justification: We explicitly cite the sources of the pre-trained weights (e.g., CLIP) and the datasets (e.g., Kinetics-400) used in our experiments.

\item {\bf New assets}
    \item[] Question: Are new assets introduced in the paper well documented and is the documentation provided alongside the assets?
    \item[] Answer: \answerYes{} % Replace by \answerYes{}, \answerNo{}, or \answerNA{}.
    \item[] Justification: The proposed commercial AI-generated video benchmark will be released with comprehensive documentation upon acceptance.

\item {\bf Crowdsourcing and research with human subjects}
    \item[] Question: For crowdsourcing experiments and research with human subjects, does the paper include the full text of instructions given to participants and screenshots, if applicable, as well as details about compensation (if any)? 
    \item[] Answer: \answerNA{} % Replace by \answerYes{}, \answerNo{}, or \answerNA{}.
    \item[] Justification: Our research does not involve crowdsourcing or human subjects.

\item {\bf Institutional review board (IRB) approvals or equivalent for research with human subjects}
    \item[] Question: Does the paper describe potential risks incurred by study participants, whether such risks were disclosed to the subjects, and whether Institutional Review Board (IRB) approvals (or an equivalent approval/review based on the requirements of your country or institution) were obtained?
    \item[] Answer: \answerNA{} % Replace by \answerYes{}, \answerNo{}, or \answerNA{}.
    \item[] Justification: Our research does not involve human subjects, making IRB approval inapplicable.

\item {\bf Declaration of LLM usage}
    \item[] Question: Does the paper describe the usage of LLMs if it is an important, original, or non-standard component of the core methods in this research? Note that if the LLM is used only for writing, editing, or formatting purposes and does \emph{not} impact the core methodology, scientific rigor, or originality of the research, declaration is not required.
    %this research? 
    \item[] Answer: \answerYes{} % Replace by \answerYes{}, \answerNo{}, or \answerNA{}.
    \item[] Justification: We clearly declared in Section 4.1 and Appendix D that we utilized Gemini-2.5 Flash to generate descriptive prompts for our dataset construction pipeline.

\end{enumerate}

\end{document}